# ESimCSE无监督对比学习联合UDA半监督学习的大标签体系文本分类模型

阮禄，周航成，冉猛，赵进，秦蛟禹，危枫，王晨子

（中国电信集团有限公司重庆分公司，重庆401147）

**摘 要**：在自然语言处理任务中，大标签体系文本分类所面临的挑战，包括标签体系多、数据分布不均衡、噪声大等问题。为了解决这些问题，通过模型中使用联合训练技术，将ESimCSE无监督对比学习和UDA半监督对比学习模型结合起来。ESimCSE模型利用无标记数据高效学习文本向量表示，以实现更好的分类效果，而UDA通过半监督学习方法利用未标记的数据进行训练，以提高模型的预测性能和稳定性，并进一步提高模型的泛化能力。此外，在模型训练过程中使用对抗训练技术FGM和PGD来提高模型的鲁棒性和可靠性。实验结果表明，在公开数据集Ruesters以及业务数据集上相对于Baseline分别有8%和10%的准确率提升，并且在业务数据集上人工验证准确率可达到15%的提升，表明该方法是有效的。

关键词： ESimCSE；UDA；半监督；无监督；对比学习；对抗训练

Ruan Lu, Zhou HangCheng, Ran Meng, Zhao Jin, Qin JiaoYu, Wei Feng, Wang ChenZi

China Telecom Corporation Limited Chongqing Branch, Chongqing 401147, China

**Abstract:** The challenges faced by text classification with large tag systems in natural language processing tasks include multiple tag systems, uneven data distribution, and high noise. To address these problems, the ESimCSE unsupervised comparative learning and UDA semi-supervised comparative learning models are combined through the use of joint training techniques in the models.The ESimCSE model efficiently learns text vector representations using unlabeled data to achieve better classification results, while UDA is trained using unlabeled data through semi-supervised learning methods to improve the prediction performance of the models and stability, and further improve the generalization ability of the model. In addition, adversarial training techniques FGM and PGD are used in the model training process to improve the robustness and reliability of the model. The experimental results show that there is an 8% and 10% accuracy improvement relative to Baseline on the public dataset Ruesters as well as on the operational dataset, respectively, and a 15% improvement in manual validation accuracy can be achieved on the operational dataset, indicating that the method is effective.

# 1 介绍

在大标签体系文本分类中，常用的是数级标签分类，数级标签通常将标签组织成一棵树，其中每个标签都有一个父节点或多个子节点。根节点代表最高级别的概念，而叶节点则表示最具体的概念。在分类过程中，每个文本实例被分配到树中的一个叶节点中。这种多级别分类方法可以更好地捕捉文本实例的语义信息，并且可以处理标签数量庞大的问题。树标签文本分类在实际应用中得到了广泛的应用。在电子商务领域，树标签分类被用于商品分类和推荐系统中。在新闻分类中，树标签分类被用于对新闻进行分类和归档。在医学领域，树标签分类被用于对医学文本进行分类和病例管理。在客服领域，树标签被应用对工单归档和标记。总之，树标签文本分类技术在许多领域中都有着广泛的应用和潜在的发展前景。尽管树标签文本分类在许多领域中得到了广泛的应用和成功，但仍然存在一些问题需要解决，包括以下几点：

**标签层次结构的不确定性**：大标签体系文本分类中由于其标签量巨大，通常使用过程都是将其划分为层次结构分类或者多标签分类，但是这种结构的不确定性会导致分类器的性能下降。在实际应用场景中，标签的层次结构非常复杂，这些标签会随着时间的推移和业务的表动而改变。

**标签不平衡问题**：在大标签体系文本分类中，有些标签可能比其他标签更常见或更重要，从而出现标签分布不平衡的情况。这可能导致机器学习算法倾向于预测常见或重要的标签，而对其他标签的分类效果较差。因此需要采取措施来解决标签不平衡问题。

**标签相似问题**：在树标签文本分类中，有些标签之间可能存在相似，例如树上同一条路径下来的两个叶子节点标签。这可能导致机器学习算法产生混淆或错误的分类结果。因此，需要采取措施来解决标签冲突问题。

在大标签体系文本分类中，当前的研究中大标签体系文本分类的算法通常分为两类：基于规则的方法和基于机器学习的方法。基于规则的方法通常是通过手工设计一些规则和特征来实现分类。而基于机器学习的方法则是利用机器学习算法自动学习文本的特征和规则。在机器学习方法中，常见的算法包括朴素贝叶斯分类器、支持向量机、决策树、神经网络（Minaee et.al & Li et.al）等。针对上述问题，本文的贡献如下：

1. 针对标签层次结构的不确定性问题，我们引入对抗干扰来增强模型鲁棒性和泛化能力，对抗干扰的核心思想是通过引入对抗样本，使得模型能够更好地处理对抗攻击。
2. 针对标签分布不平衡问题，我们通过集成半监督学习 UDA 和 mixup 数据增强到模型训练中去，同时训练监督数据和无监督训练来提升模型的认知能力。
3. 针对标签相似问题，我们引入无监督对比学习 ESimCSE 模型，通过无监督对比学习的方式学习样本向量嵌入之间的关系，最终达到一个更准确的分类结果。

针对这些优化方案，我们在开源数据集 Reuters 以及我们的客服领域工单数据集中（后续称为业务数据集）都得到了不错的效果，其中我们在 Reuters 的结果是 98%，在业务数据中我们模型在原始标签上的评估上面准确率是 80%，但是根据业务实际检测后准确率达到了 94%进一步证明了我们的方法不仅可以提升基准模型的准确率，同时可以纠正一些分类错误的数据。

# 2 相关工作

## 2.1 对抗训练

对抗训练常用于对 CV 领域的任务中处理中，它通常被用于生成对于人类来说 看起来"几乎一样，但对于模型来说预测结果却完全不一样的样本，神经网络由于其线性的特点，

很容易受到线性扰动的攻击，对抗攻击可以暴露机器学习模型的脆弱性，进而提高模型的鲁棒性和可解释性。Szegedy 等人.2014 和 Goodfellow 等人.2015 提供了一种快速生成对抗样本的方法 FGSM。它们不是搜索给定数据点的整个邻域，而是直接沿着用于训练模型的损失函数（相对于输入）的梯度方向移动以达到对抗的目的。Miyato 等人.2016 通过在不查看标签信息的情况下计算对抗方向，进一步改进了这一想法。

然而 NLP 任务模型输入通常由单词（有时是字符）组成，它们本质上形成了一个离散的空间，其欧式距离恒为$\sqrt{2}$，因此理论上不存在小扰动。Miyato 等人.2017 利用书面文字通常通过嵌入映射到连续空间这一事实，对 FGSM 中计算扰动的部分做了一点简单的修改。假设输入文本序列的 Embedding vectors$[v_1, v_2, ..., v_T]$为$x$，通过对嵌入进行$\Delta x = \epsilon \cdot \frac{g}{\|g\|_2}$对抗性扰动，其中$g = \nabla_x L(x, y; \theta)$，该方法称为 FGM，并证明了其对抗性和虚拟对抗性训练适用于文本。他们在各种文本分类任务上取得了效果，并学习了更高质量的词嵌入。Jia 等人通过在斯坦福问答数据集 (SQuAD) 中插入句子来干扰阅读理解模型，这些句子添加到文本中而不改变所询问的信息。Glockner 等人用同义词或反义词替换 SNLI 测试集中的单个单词，这导致与原始测试集相比模型性能显著下降。Belinkov 等人通过引入对人类没有挑战的拼写错误或字符交换等噪声来打破基于字符的机器翻译。Sato 等人。现有词嵌入的方向上扰动嵌入，并用梯度信息对这些方向进行加权。这种微妙的修改允许轻松重建可能欺骗模型的真实单词。

现有的常用的文本对抗 Madry 等人在 2018 年总结了之前的工作并提出了 Min-Max 公式，对抗训练可以写成如下格式：

$$\min_{\theta} \mathbb{E}_{(x,y) \sim \mathcal{D}} \left[ \max_{\Delta x \in \Omega} L(x + \Delta x, y; \theta) \right]$$

其中$\mathcal{D}$代表数据集，$x$代表输入，$y$代表标签，$\theta$是模型参数，$L(x + \Delta x, y; \theta)$是单个样本的$loss$，$\Delta x$是扰动，$\Omega$是扰动空间。这个式子可以分步理解如下：

1.往$x$里注入扰动$\Delta x$，$\Delta x$的目标是让$L(x + \Delta x, y; \theta)$越大越好，也就是极可能让模型在预测过程中犯错，当然为了看控制这种扰动，$\Delta x$也要满足一定的约束，常规的约束一般是$\|\Delta x\| \leq \epsilon$，其中$\epsilon$是一个常数。

2.每个样本都构造出对抗样本$x + \Delta x$后，用$(x + \Delta x, y)$作为数据取最小化损失$loss$来进行梯度下降更新模型参数$\theta$，反复重复这些步骤直到模型收敛。

并且 Madry 等人提出了 FGM 的优化方法 PGD，FGM 的思路是梯度上升，通过一步到位式的走到最佳约束点，PGD 通过小步走，多走几步，最终走出一个$\epsilon$的扰动空间重新映射回球面上，从而保证干扰不要过大，具体公式是$x_{t+1} = \prod_{x+S} \left( x_t + \alpha \frac{g(x_t)}{\|g(x_t)\|_2} \right)$，$g(x_t) = \nabla_{x_t} L(x_t, y; \theta)$，其中$S = \{r \in \mathbb{R}^d : \|r\|_2 \leq \epsilon\}$为扰动空间，$\alpha$为小步的步长。

## 2.2 半监督训练

监督学习使用标记的数据集训练机器学习模型。有标记标签通常在数据中可用，但该过程可能涉及人类专家，将标签添加到原始数据中，以向模型显示目标属性（答案）。无监督学习是指模型在没有人类监督的情况下尝试自己挖掘未标记数据中隐藏的模式、差异和相似之处。半监督学习（SSL）是一种机器学习技术，其原理是将监督学习和无监督学习技术联系起来，使用一小部分标记数据和大量未标记数据来训练预测模型。半监督学习通常针对未标注数据相关的正则化项进行设置，其通常有以下两种：

1. 熵最小化 (Entropy Minimization)：根据半监督学习的 Cluster 假设，决策边界应该尽可能地通过数据较为稀疏的地方（低密度区），从而避免把密集的样本数据点分到决策边界的两侧。也就是模型通过对未标记数据进行低熵预测，即熵最小化。

2. 一致性正则 (Consistency Regularization)： 对于未标记数据，我们希望模型在其输入受到扰动时产生相同的输出分布。即：$\sum_{i=1}^{N} \|z(x_i) - \tilde{z}(x_i)\|^2$，$z$定义为人工标签，该公式的含义是构建标一套标准来指导一致性正则的计算，$z$也可以是模型历史平均预测值或模型单轮平均预测值。$\tilde{z}$定义为预测标签，为模型当前时刻对无标注数据的预测，对其输入可进行强增强或对抗扰动。

常见的半监督方法是 Dong-Hyun Lee 等提出的 Pseudo-Label 方法。该方法的损失函数是 $L = \frac{1}{n}\sum_{m=1}^{n}\sum_{i=1}^{C} L(y_i^m, f_i^m) + \alpha(t)\frac{1}{n'}\sum_{m=1}^{n'}\sum_{i=1}^{C} L(y_i'^m, f_i'^m)$。从损失函数可知该方法通过选举每个未标注样本的最大概率作为其伪标签，然后利用熵最小化的思想，用未标注数据和伪标签进行训练来引导模型预测的类概率逼近其中一个类别，从而将伪标签条件熵减到最小。随后 Laine 等人提出了 Π-Model 模型和 Temporal Ensembling 模型，Π-Model 通过对无标注数据输入进行了两次不同的随机数据增强、并引入一致性正则到损失函数 L2 损失中（$loss \leftarrow -\frac{1}{|B|}\sum_{i \in (B \cap L)} \log z_i[y_i] + w(t)\frac{1}{C|B|}\sum_{i \in B} \|z_i - \tilde{z}_i\|^2$）以增强单次计算的噪音、同时加速训练速度。Temporal Ensembling 采用时序融合模型，引入 EMA 加权平均和，对同一个训练步骤进行数据增强，以达到半监督的目的。temporal ensembling 中每个 epoch 都要更新一次伪标签，如果面对的是很大的数据集，那么这种更新方式会变得很缓慢，从而带来这样很容易让模型陷入 confirmation bias (即错误被自己放大)等问题，为了解决这个问题，Antti Tarvainen 等人提出来了 Mean Teacher， Mean Teacher 通过对模型参数进行 EMA 平均，将平均模型作为 Teacher 预测人工标签，由当前模型（看作 Student）预测伪标签，从而解决模型自己生成标签时带来的不正确性问题。Virtual Adversarial Training（VAT）采取对抗训练的方式添加噪音。采用$p(y|x,\theta)$构建一个伪标签，并根据这个伪标签来计算对抗扰动方向（通常是梯度上升的方向，即正梯度方向），同时利用 KL 散度 $D_{kl}[p(y|x,\theta), p(y|x+r_{adv},\theta)]$来计算一致性正则。以上方法都只使用了熵最小化或者一致性正则的单一方法，Xie 等人提出了 UDA(Unsupervised Data Augmentation)方法，该方法同时引入了一致性正则和熵最小化正则。通过一致性正则的思想使用回译和非核心词替换对文本进行无监督增强，同时引入熵最小化正则对无监督信号$p(y|x,\tilde{\theta})$进行 sharpen 操作构建人工标签，将人工标签于强增强后的预测标签共同构建一致性正则，通过 KL 散度 $\mathcal{D}(p_\theta(y|x) \| p_\theta(y|x,\epsilon))$计算损失，同时在计算过程中进行了 confidence-based masking，将低于置信度阈值剔除去，不参与 loss 计算以减少低置信度对模型的影响，同时采用训练信号退火(TSA)方法防止对标注数据过拟合。

## 2.3 Esimcse 句向量表示

BERT 等语言模型在多数 NLP 任务中取得优异的表现，但如果直接取 BERT 输出的句向量作表征，取得的效果甚至还不如 Glove 词向量。Bohan Li 等人的 Bert-flow 中指出，产生该现象的原因是 BERT 模型的各向异性过高，Transformer 模型的输出中高频词汇分布集中，低频词汇分布分散，整个向量空间类似于锥形结构。余弦相似度使用的前提是向量空间在标准正交基下，而 BERT 输出的句向量很明显 不符合该条件，因此直接使用 BERT 输出的句向量计算余弦相似度，效果表现很差。针对该问题 Reimers 等人提出的 Sentence-Bert 采用孪生网络的双塔结构，将推理的次数降低到了 $O(n)$，但计算相似度的点积复杂度还是 $O(n^2)$，Bert-flow 引入 flow 流式模型，使得输出映射具有同向性，将 BERT 句向量映射到标准高斯空间（Gaussian space）。随后 Jianlin Su 等人提出的 BERT-Whitening 在句向量维度，通过一个白化的操作直接校正局向量的协方差矩阵 $W = U\sqrt{\Lambda^{-1}}$，从而增强增强句子表示的

各向同性，同时白化操作还能降低句子表示的维度降低句子存储成本加速模型检索速度。Yuanmeng Yan 等人提出的 ConSERT 引入对比学习（Contrastive Learning），通过对比学习 $f = \text{Concat}(r_1, r_2, |r_1 - r_2|)$ 拉近相同样本的距离、拉远不同样本的距离，来刻画样本本身的表示，以解决 BERT 表示的塌缩问题。Yianyu Gao 等人提出的 SimCSE 方法包含了无监督和有监督的两种方法。无监督方法采用 dropout 技术，对原始文本进行数据增强，从而构造出正样本，用于后续对比学习训练；监督学习方法借助于文本蕴含（自然语言推理）数据集，将蕴涵对作为正例，矛盾对作为负例，用于后续对比学习训练。并且通过对比学习解决了预训练 Embedding 的各向异性问题，使其空间分布更均匀，当有监督数据可用时，可以使正样本直接更紧密。但是 SimCSE 存在两个问题：第一个问题是正例的构建方法有待提升，基于 dropout mask 构造正例的方法会使模型倾向于将长度相近的语句认定为相似。第二个问题是负例的构建方法有待提升。因为 SimCSE 是基于负例的对比学习方法，理论上 batch 内负例越多模型学习的效果越好，但是 batch size 设置过大会导致性能下降。鉴于上述问题，Xing Wu 等人提出了 ESimCSE 方法，该方法通过引入重复词正向句子构架和动量对比优化负例构建来优化 SimCSE，通过词重复的方式构建正例，不仅可以很好的保持文本语义相似度，而且还能缓解语句长度差异对模型效果的影响。为了有效的扩展负例对，同时缓解扩展 batch size 时 GPU 内存的限制，通过动量对比(momentum contrast)的方式，维护一个固定大小的队列，从而可以重用邻近 batch 中编码句子的 embedding。从而可以更充分的学习。

## 3 我们的方法

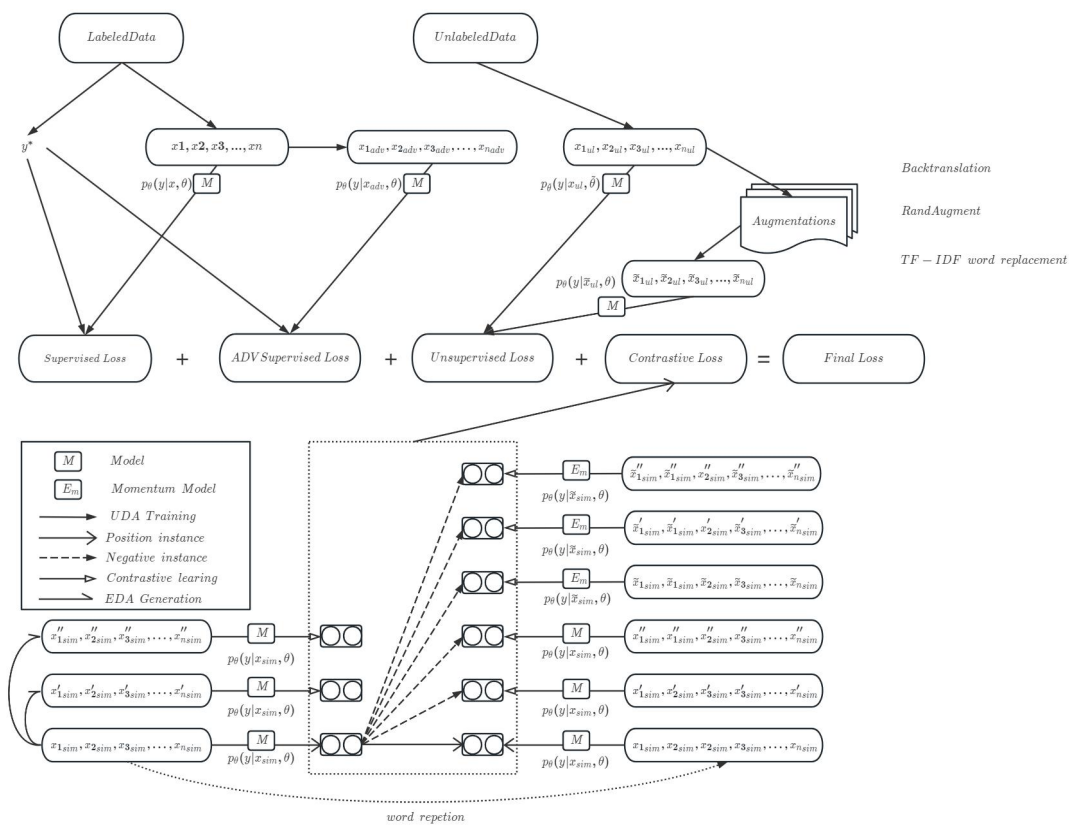

**图 1 整体模型框架图**

本文中提出了一种双重半监督训练的分类训练方法，具体的我们通过引入 google 半监督训练方法 UDA 来对数据增强并进行半监督训练以解决多标签文本分类任务中的文本不均衡问题，同时在 UDA 半监督训练范式中我们也引入了对抗训练来提升模型的鲁棒性和泛华

能力，避免模型的过拟合。除了 UDA 半监督训练之外，我们也引入了半监督对比学习文本编码模型 ESimCSE 模型来训练模型，以解决树标签文本分类任务中的底层节点文本特征相似度高导致的模型模型分类能力下降的问题。

### 3.1 UDA 半监督训练

为了使用 uda 半监督训练方法，对于输入的单条数据点 $D = (x_1, x_2, x_3, ..., x_n), y^*$,其中 $x_i$ 表示训练数据中离散化的单个特征值，$y^*$ 表示有监督标注的标签。我们的目标是学习一个共享分类器 $M = p_\theta(y|x, \theta)$，来根据 $x$ 预测真实标签 $y^*$，并计算监督损失 $Supvised\ Loss\ L_{sup}$，其中 $\theta$ 表示模型参数。同时我们使用对抗训练方法针对单次输入对输入 $x$ 进行扰动得到 $x_{adv}$，然后使用共享模型 $p_\theta(y|x_{adv}, \theta)$ 来预测 $y$，并分别预测值和真实值计算出一个对抗损失 $ADV\ Supervised\ Loss\ L_{adv}$，对于无标签的数据 $x_{ul}$,我们使用数据增强方法计算出一个增强后的新数据 $\tilde{x}_{ul}$，并分别用模型来预测并计算无监督训练部分的损失 $Unsupervised\ Loss\ L_{Unsup}$。

对于有监督训练部分，在每一轮训练中我们按照标准的有监督训练流程先求解一个预测值 $f^*(x) = p_\theta(y|x, \theta)$,然后使用交叉熵损失的计算方法来计算有监督训练部分的损失：

$$L_{Sup} = E_{x \sim p_L(x)}[-\log p_\theta(f^*(x)|x)]$$

对无监督训练部分，我们使用回译（Back translation）、TF-DF 词重复等方法将原始未标注的数据 $x_{ul}$ 增强为 $\tilde{x}_{ul}$，对于原始未标注数据 $x_{ul}$ 我们使用当前参数 θ 的备份 $\tilde{\theta}$ 来计算一个新的预测值 $f^*(x_{ul}) = p_{\tilde{\theta}}(y|x_{ul})$，同时对于增强后的数据 $\tilde{x}_{ul}$，我么使用当前参数 θ 来计算一个输出 $f^*(x_{ul}) = p_\theta(y|x_{ul})$，然后使用 KL 散度来衡量两个预测值分布之间的差异性 $\tilde{x}_{ul}$ 作为一致性损失。最终无监督部分的损失为：

$$L_{Unsup} = E_{x_{ul} \sim p_U(x_{ul})} E_{\tilde{x}_{ul} \sim q(\tilde{x}_{ul}|x_{ul})}[KL(p_{\tilde{\theta}}(y|x_{ul}) \| p_\theta(y|\tilde{x}_{ul}))]$$

$$KL(p_{\tilde{\theta}}(y|x_{ul}) \| p_\theta(y|\tilde{x}_{ul})) = \sum_{i=1}^{n} p_{x_{ul}} \log\left(\frac{p_{(x_{ul})}}{q_{(\tilde{x}_{ul})}}\right) = \sum_{i=1}^{n} p_{x_{ul}} \log(p_{x_{ul}}) - \sum_{i=1}^{n} p_{\tilde{x}_{ul}} \log(\tilde{x}_{ul})$$

对于对抗训练部分，由于对抗攻击的搜索的空间巨大，带标签的基本数据有限，我们对原有数据在嵌入空间上进行离散对抗攻击增强，以改善共享模型 $M = p_\theta(y|x, \theta)$ 的泛化能力和鲁棒性，具体的我们定义对抗训练部分的损失为：

$$L_{adv} = E_{x_{adv} \sim p_L(x)}[-\log p_\theta(f^*(x_{adv})|x_{adv})]$$

将 $x_{adv}$ 展开可写成：

$$L_{adv} = E_{x_{adv} \sim p_L(x)}[-\log p_\theta(f^*(x + \Delta x)|x + \Delta x)]$$

其中 $\Delta x$ 为对抗上面的扰动，我们使用以 $\epsilon$ 为界的 Frobenius 约束 $\Delta x$，然后通过投影梯度下降算法 PGD 对上诉公式进行求解，PGD 是大规模约束优化的标准方法，每次迭代的步长 $\delta$ 为：

$$\delta_{l+1} = \Pi_{\|\delta\| \leq \epsilon}(\delta_t + \eta g(\delta_L)/\|g(\delta_l)\|_F)$$

为了更直观的显示整个对抗训练的步骤我们提供了对抗训练损失求解的算法步骤细节：

表 1 对抗训练算法步骤表

```
Algorithm 1: 对抗训练算法步骤
Require: Traing samples D, perturbation bound ϵ, Weighting of the anti-training component
α, acent steps K, acent steps size ξ, the total number of epochs T, the cariance of the random
initialization σ²
01: Initialize θ
02: for epoch = 1…T do
03:     for each batch ξ from D do
04:         δ ~ 𝒩(0, σ²I)
05:         g₀^θ ← 0
06:         for t = 1…K do
07:             calculate gradient of x_t:
08:             x_{t+1}^θ ← x_t^θ + (1/K)𝔼_B[∇_θ L(x_t, y, θ)]
09:             update the perturbation δ by gradient ascend:
10:             g_{adv}^δ ← ∇_δ L_{adv}
11:             δ_t ← Π_{‖δ‖_F ≤ ϵ}(δ_{t-1} + η · g_{adv}^δ/‖g_{adv}^δ‖_F)
12:         end for
13:         L_{adv} ← αL_{adv}
14:     end for
15: end for
```

最终的 UDA 部分的损失可统一写成：$L_{uda} = L_{sup} + L_{Unsup} + L_{adv}$，完整的求解目标可展开为：

$$L_{uda} = E_{x \sim p_L(x)}[-\log p_\theta(f^*(x)|x)] + \lambda E_{x_{ul} \sim p_U(x_{ul})} E_{\tilde{x}_{ul} \sim q(\tilde{x}_{ul}|x_{ul})}[KL(p_{\hat\theta}(y|x_{ul}) \| p_\theta(y|\tilde{x}_{ul}))]$$
$$+ \alpha E_{x_{adv} \sim p_L(x)}[-\log p_\theta(f^*(x_{adv})|x_{adv})]$$

其中 $KL = \sum_{i=1}^{n} p_{x_{ul}} \log\left(\frac{p_{(x_{ul})}}{q_{(\tilde{x}_{ul})}}\right) = \sum_{i=1}^{n} p_{x_{ul}} \log(p_{x_{ul}}) - \sum_{i=1}^{n} p_{\tilde{x}_{ul}} \log(\tilde{x}_{ul})$ 表示 KL 散度，

用来衡量两个分布之间的差异，原始论文中使用的是交叉熵 $CE = -\sum_{i=1}^{n} p_{x_i} \log(q_{x_i})$ 来衡量，这里由于我们模型和 CE 损失的适配结果不是很理想我们最终选用 KL 来衡量 $q(\hat{x}|x)$ 是数据强转换，$\hat{\theta}$ 是当前参数 $\theta$ 的备份副本，这里我们只使用备份参数来计算，但并不会更新 $\hat{\theta}$ 参数，$\lambda$ 表示无监督部分训练损失的权重，以防止无监督训练部分的损失过大带来的有监督部分损失失效等问题。实际实验中我们遵循 VAT 模型建议的那样，将 $\lambda$ 的值设置为 1。在每次迭代中我们在一个小批量的样本中计算监督损失，并在一个小批量的无监督数据上使用过 KL 散度来计算无监督部分的一致性损失。$\alpha$ 表示对抗训练部分的权重，这里对抗训练我们作为一个正则项来使用，用来矫正和提升模型的泛化能力。

### 3.2 ESIMCSE 半监督对比学习

为了解决子树上面叶子节点数值高相似度的问题，我们引入了无监督动量对比学习 ESimCSE 模型来对句子进行编码。具体的，按照 ESimCSE 计算流程的要求我们使用 EDA 算法来构建输入文本的负例（negative input），对于输入无标签样本 $D = \{x_{1_{sim}}, x_{2_{sim}}, x_{3_{sim}}, ..., x_{n_{sim}}\}$，我们使用 EDA 算法中的随机插入、反义词替换、随机删除等方法随机的对原始文本进行改造成 $\{x'_{1_{sim}}, x'_{1_{sim}}, x'_{2_{sim}}, x'_{3_{sim}}, ..., x'_{n_{sim}}\}$、$\{x''_{1_{sim}}, x''_{1_{sim}}, x''_{2_{sim}}, x''_{3_{sim}}, ..., x''_{n_{sim}}\}$ 等，之后我们通过编辑距离（edit distance）来粗略的估算构造的句子对之间的相似度，然后将相似度取整后得到文本对间的相关性标签形成负例文本对：

$$\{[(x_{1_{sim}}, x_{2_{sim}}, x_{3_{sim}}, ..., x_{n_{sim}}), (x'_{1_{sim}}, x'_{1_{sim}}, x'_{2_{sim}}, x'_{3_{sim}}, ..., x'_{n_{sim}})], [y^{neg}]\}$$
$$\{[(x_{1_{sim}}, x_{2_{sim}}, x_{3_{sim}}, ..., x_{n_{sim}}), (x''_{1_{sim}}, x''_{1_{sim}}, x''_{2_{sim}}, x''_{3_{sim}}, ..., x''_{n_{sim}})], [y^{neg}]\}$$

对于正例(positive input)的构造，我们遵循原始论文的要求使用词重复（word repetition）的方式构建正例，首先我们定义重复标记的数量为：$dup_{len} \in [0, max(2, int(dup_{rate} * N))]$，其中 $dup_{rate}$ 表示最大重复了，然后我们在扩展序列中随机抽取一个长度为 $dup_{len}$ 的集合中的一个字词来作为重复词构建文本对：

$$\{[(x_{1_{sim}}, x_{2_{sim}}, x_{3_{sim}}, ..., x_{n_{sim}}), (x'_{1_{sim}}, x'_{1_{sim}}, x'_{2_{sim}}, x'_{3_{sim}}, ..., x'_{n_{sim}})], [y^{pos}]\}$$
$$\{[(x_{1_{sim}}, x_{2_{sim}}, x_{3_{sim}}, ..., x_{n_{sim}}), (x''_{1_{sim}}, x''_{1_{sim}}, x''_{2_{sim}}, x''_{3_{sim}}, ..., x''_{n_{sim}})], [y^{pos}]\}$$

为了计算 ESimCSE 无监督动量对比学习的损失，对于任意一个文本对$\{x_{sim}, x^+_{sim}\}$，我们分别将文本对间的单个文本输入模型进行编码，利用 Transformers 中的全连接层和注意力概率上的 dropout 掩码，并输出两个单独的文本嵌入来构建正对嵌入表示，具体处理如下所示：

$$h_i = p_\theta(x_{sim}, z_{sim}), h_i^+ = p_\theta(x^+_{sim}, z^+_{sim})$$

在批量大小为 N 的 mini-batch 中，每个句子都有$h_i$和$h_i^+$，根据对比学习的学习目标$w.r.t$的要求我们初步定义损失函数 Contrastive Loss 如下：

$$L_{con} = -log\frac{e^{sim(h_i, h_i^+)/\tau}}{\sum_{j=1}^{N} e^{sim(h_i, h_j^+)/\tau}}$$

其中$\tau$是一个温度超参数，$sim(h_i, h_i^+)$是一个相似度度量函数，用来衡量两个输出文本嵌入向量之间的相似性，这里比较常用的一般是余弦相似度函数。在对比学习中，理论上负例越多，文本对之间的比较效果会更好（chen et al .,2020）。这里我们通过维护一个特殊的队列来扩展和记录负例，按照先进先出的思想，该队列总是对当前批次的句子进行排队，同时将最老的编码嵌入出队，目的是重用前面批次中的编码嵌入。具体的，我们将原始共享模型的参数记为$\theta$，动量更新的编码器的参数记为$\theta_m$，对$\theta_m$的更新步骤如下：

$$\theta_m \leftarrow \gamma\theta_m + (1-\gamma)\theta$$

这里$\gamma \in [0,1)$是动量参数的系数，每一轮反向更新参数的时候我们只更新当前的参数$\theta$
为了训练，引入对量对比之后，无监督对比学习的损失函数可以进一步写成：

$$L_{con} = -log\frac{e^{sim(h_i, h_i^+)/\tau}}{\sum_{j=1}^{N} e^{sim(h_i, h_j^+)/\tau} + \sum_{m=1}^{M} e^{sim(h_i, h_m^+)/\tau}}$$

其中$h_m^+$表示动量更新队列中嵌入的句子，$M$是队列大小。

最终我们定义的模型损失 *Final Loss* 如下：
$$L = \omega L_{uda} + (1-\omega)L_{con}$$
其中$\omega$是整个半监督训练中控制两者损失大小权重的一个超参数，训练过程中，我们使用有标签的数据和没有标签的数据分别去训练 UDA 半监督分类模型和无监督编码模型 ESimCSE，并分别计算两次计算中的损失，我们底层使用的是 roberta 作为基础模型。

## 4 实验

### 4.1 数据集

为了验证模型的分类能力，我们选择了两个分类类目比较多，并且分类类目比较复杂的数据集：Ruesters 数据集和我们业务中实际的数据集。其中 Reuters 全称 Reuters-21578 ，是一个用于文档分类的基准多标签数据集，它由 1987 年从路透社新闻专线中提取的新闻内容组成。在这个数据集中，每个样本都被标记为总共 90 个新闻类别中的一个或多个新闻类别。我们使用标准的 ModApte 拆分 并只使用了其中的第一个标签来进行实验；业务数据集是我们实际投诉工单中的业务数据，该数据的文本内容是客户对于实际使用的产品的具体投诉内容的描述，该内容包含了投诉事件的时间、起因、经过、处理结果、客服最终回复等详细内容，其对应的标签体系是一个数级路径标签，最终对应到叶子节点上的标签总共有 300 个标签，该数据的特点一是标签路径中的前几级标签相等的越多，其最终的内容相似

性越大，特点二是数据是由不同的人打标的，由于标签体系众多且相似度大，同一条数据可能不同的人会标注出不通的标签，这会导致数据有一定的业务错误率 BER（Business Error Rate）。下图 2 是我们对两个数据集的词数量和样本对应关系的一个分布情况，下表 1 是我们实际训练中使用得两个数据集的具体划分信息，其中 x 坐标轴是文本长度，y 坐标轴是对应长度文本的数量，可以看到 Reuters 数据集长度集中在 250 以下，业务数据集的长度集中在 100 和 400 这两个长度左右，相比于 Reuters 数据集，业务数据集分布较为均衡一些。

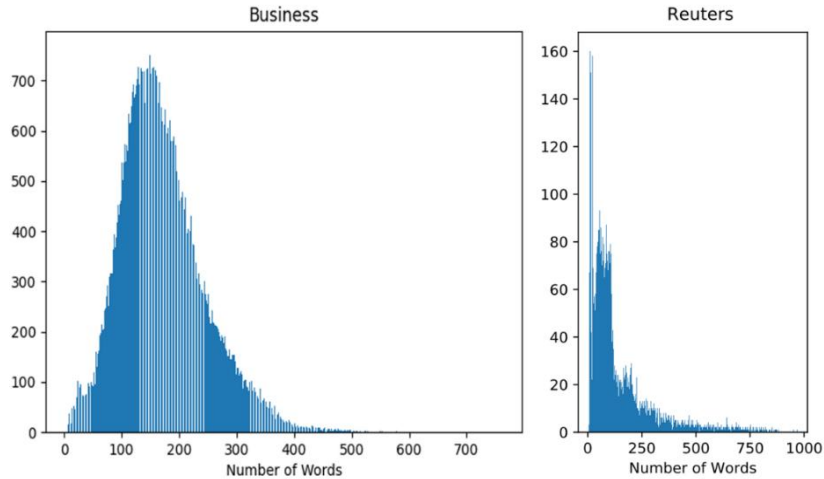

图 2 数据分布信息图

Table 2 Datasets Information

| 参数 | Reuters | Business |
|---|---|---|
| 总类别数 | 90 | 300 |
| 有标注样本数 | 6W | 30W |
| 无标注的样本数（UDA） | 1W | 6W |
| 无标注的样本数（ESimCSE） | 8K | 3W |
| 每批无标注样本数（UDA） | 24 | 64 |
| 每批无标注样本数（ESimCSE） | 24 | 32 |
| 测试样本数 | 8K | 3W |
| RA样本数 | 5K | 2W |
| **SA样本数** | **5K** | **2W** |

表 2 数据集划分情况标

## 4.2 实验结果

为了对比最终模型的效果我们选取了 10 个组合模型。其中 Baseline 模型是用的底层是 Roberta 原始权重来训练的一个有监督分类模型，该模型只使用了上述表 2 中的有标注部分的数据集进行训练；Baseline+PGD 和 Baseline+FGM 模型是在原始 Baseline 模型的基础上引入了 FGM 对抗训练和 PGD 对抗训练，同样这两个模型也是使用了表 2 中的有标注样本进行训练；UDA Baseline 模型是在原始模型的基础引入 UDA 半监督训练方法，我们使用的训练数据集是有标注样本和无标注样本（UDA）；Baseline+ESimCSE 是在原始 Baseline 模型的基础上引入了 ESimCSE 无监督对比学习，两者使用的基础模型都是 Roberta 模型，使用的无标签数据是无标注样本（ESimCSE）；UDA Baseline+ESimCSe 是同时引入了半监督学习和无监督编码学习算法；UDA Baseline+PGD、UDA Baseline+FGM、UDA Baseline +ESimCSE+ FGM 以及 UDA Baseline +ESimCSE+ PGD 等方法都是在上述模型的基础上面加入对抗训练。

对于两个数据集我们都设定了准确率（Accuracy）和 F1 指标，对于我们实际业务中的数据，

我们还增加了一个评估指标，人工验证指标准确率（Manual verify Acc），该准确率的指标是只使用单个人来验证模型最终的分类能力，以验证模型对错误率 BER 的泛化情况。具体情况如表 3 所示，实验结果表明在引入对抗训练之后，实验结果有好有坏，但整体而言 PGM 对抗训练的效果明显要优于 FGM 对抗训练的实际效果；在引入 UDA 半监督对抗训练之后，UDA 在业务数据集上面的提升空间比较明显，原因是业务数据中使用的无标签数据很多，并且相比于 Reusters 数据集，业务数据集数据更平衡一些，证明了半监督训练学习到了未标注数据的样本特征，进一步证明了 UDA 半监督的有效性；在引入 ESimCSE 模型之后，结果如图中加粗的数据所示，在 Reusters 数据集和业务数据集上模型准确率（Accrary）和 F1 指标都大幅上升，并且通过人工验证后，实际准确率还会提升 5%，说明了模型一定程度上可以纠正原有错误标注的数据，从侧面反应了模型的泛化能力。

表 3 模型 Accuracy & F1 测试结果表

Table 3 Model Accuracy & F1 Test Results

| Methods | Reuters | | Business | | |
|---|---|---|---|---|---|
| | Accyrary | Micro F1 | Accyrary | Micro F1 | Manual verifi Acc |
| Baseline(Roberta weights) | 81.1 | 75.3 | 68.3 | 54.9 | 67.8 |
| Baseline + FGM | 79.2 | 76.9 | 68.7 | 55.1 | 68.9 |
| Baseline + PGD | 81.4 | 77.2 | 69.1 | 55.6 | 70.1 |
| UDA Baseline(Roberta weights) | 83.7 | 79.1 | 74.5 | 60.2 | 80.5 |
| Baseline + ESimCSE(Roberta weights) | 85.3 | 80.5 | 72.8 | 59.3 | 77.4 |
| UDA baseline + FGM | 83.9 | 80.3 | 74.3 | 59.8 | 76.2 |
| UDA Baseline + PGD | 84.6 | 81.7 | 71.4 | 58.4 | 76.9 |
| UDA Baseline + ESimCSE | **86.0±0.22** | **83.0±0.14** | **76.0±0.31** | **62.0±0.12** | **81.0±0.22** |
| UDA Baseline + ESimCSE + FGM | 88.0±0.70 | 85.0±0.46 | 77.0±0.59 | 62.0±0.68 | 82.0±0.58 |
| UDA Baseline + ESimCSE + PGD | **89.0±0.53** | **85.0±0.70** | **78.0±0.27** | **63.0±0.42** | **83.0±0.61** |

为了直观地验证不同优化技巧下模型之间的差异，我们使用了相同的输入并经过网络结构，得到了最终输出（logits 层，即 softmax 层之前的输出）。我们使用 t-SNE 降维技术将输出向量降维为 $1*64$ 的形状，并对每个模型的向量进行了比较，结果如图 3 所示。结果表明，原始基准模型的不同类别特征值之间的差异较小，说明模型没有学习到特征之间的差异性。与未加对抗训练的模型相比，加入对抗训练的模型输出向量的特征值之间的取值空间较大，不同位置之间的特征值差异性变大，使模型更容易区分具体的类别。这进一步证明了对抗训练可以提高模型的泛化能力和鲁棒性。

使用 UDA 无监督训练和 ESimCSE 训练后的输出更加合理。尽管对抗训练可以通过扰动来增加模型输出之间的差异性，但特征值之间的取值空间太大会导致模型预测结果的不确定性增加，并且在训练过程中可能出现梯度值过大的情况，导致梯度消失或梯度爆炸等问题。然而，当模型与 ESimCSE 一起训练时，模型更加关注于编码任务，将模型输出向量的特征值控制在合理的区间内。

向量化对比图：

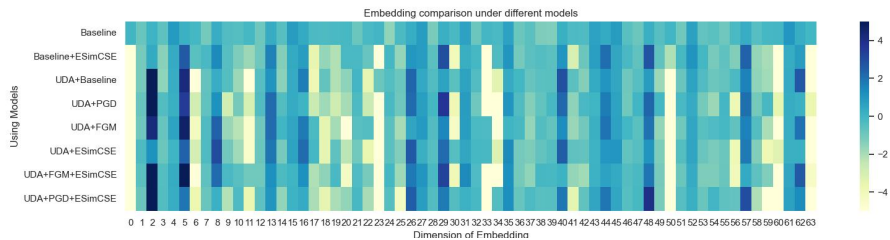

图 3 输出向量对比可视化图

本次实验中，我们针对 Ruesters 和业务数据集分别训练了几个模型，并对比了它们的损失下降情况如下图 4 和下图 5 所示。通过对两个数据集上面的损失下降图的观察，我们发现 ESimCSE+UDA+PGD 的模型损失下降较为平稳，这可能是因为加入了对抗训练和无监督数据增强技术，使得模型更具有鲁棒性和泛化能力。而 UDA 和 Baseline 的损失下降速度较慢且比较平稳，这可能是因为模型没有引入对抗训练和数据增强技术，导致模型学习速度较

慢。另外，我们发现加入 PGD 对抗训练的模型损失下降过程中存在较大的震动，这可能是

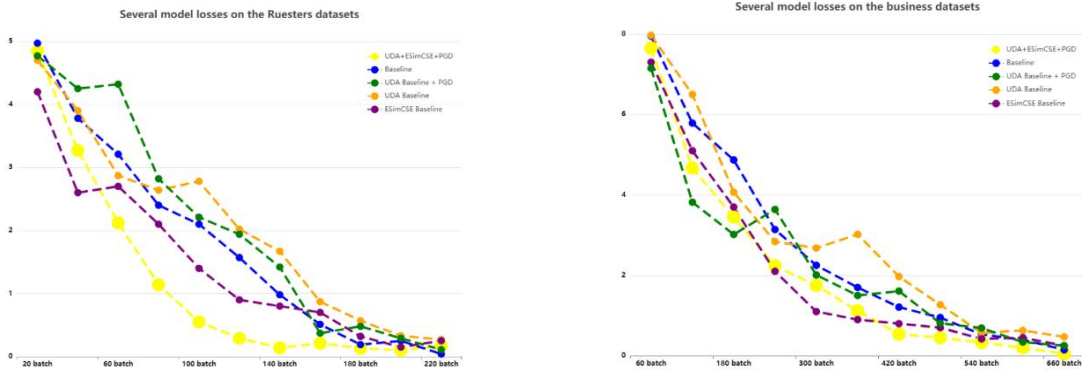

因为对抗训练增加了噪声，导致模型难以收敛从而出现损失曲线不平滑的情况。综上所述，我们可以发现，在训练过程中，对抗训练和数据增强技术可以提升模型性能，但也需要注意对训练过程的影响，以避免出现不良的影响。损失下降图：

**图 4 损失下降图**

性能影响：

为了准确评价对抗训练对模型整体性能的影响，我们定义了两个测试指标：标准正确率（Standard Accurary，SA）和鲁棒正确率（Robustness Accurary，RA），SA 表示随机抽取的正常测试样本上的模型准确率，RA 表示随机抽取测试样本经过对抗干扰的之后的数据，对于两个数据集而言我们使用的 SA 和 RA 上面的测试数据情况图表 2 所示。对于 Baseline、UDA Baseline 以及以及 UDA Baseline + ESimCSE 模型我们分别选取了 FGM 对抗后的模型和 FGM 对抗训练的模型，对于使用 FGM 训练的模型我们生成样本使用标准 PGD 公式生成 RA 数据集：

$$x_{t+1} = \prod_{x+s}\left(x_t + \rho \frac{\nabla_x L(x_t, y, \theta)}{\|\nabla_x L(x_t, y, \theta)\|_2}\right)$$

式中 $S$ 是约束空间，$\alpha$ 为步长，对于用 PGD 训练的模型我们生成样本使用标准 FGM 公式生成 RA 数据集：

$$x_t = x_t + \Delta x = x_t + \varepsilon \cdot \frac{\nabla_x L(x_t, y, \theta)}{\|\nabla_x L(x_t, y, \theta)\|_2}$$

式中 $\varepsilon$ 为缩放因子，$\nabla_x L(x_t, y, \theta)$ 为损失函数对于 $x_t$ 的偏导，即损失函数在该点的梯度。最终在几个模型上面的 RA，SA 测试结果如下表 4 所示，表中加粗的字段为较大值，就两个数据集而言，对于 6 个对抗训练模型，都有 4 个模型的 RA 指标大于 SA 指标；对于每一个数据集，有分别有 3 个 FGM 对抗训练和 3 个 PGD 对抗训练，结果显示 PGD 上面训练的 RA 值大于 SA 值的有 2 个，小于的有 1 个，两个对比实验都表明对抗训练的有效性。

**表 4 模型对抗训练验证表**

| Methods | Hyperparameters | | Reuters | | Business | |
| --- | --- | --- | --- | --- | --- | --- |
| | $\epsilon$ | $\delta$ | SA | RA | SA | RA |
| Baseline + FGM | | | **77.9** | 77.7 | 68.7 | **69.2** |
| Baseline + PGD | | | 81.7 | **82.3** | **69.1** | 68.5 |
| UDA Baseline + FGM | 2.57E-03 | 10 | **83.2** | 81.5 | 74.3 | **75.4** |
| UDA Baseline + PGD | | | 82.1 | **82.2** | 71.4 | **74.7** |
| UDA Baseline + ESimCSE + FGM | | | 87.7 | **88.1** | 77.5 | **79.1** |
| UDA Baseline + ESimCSE + PGD | | | 87.5 | **87.6** | **78.3** | 77.9 |

## 5 结束语

本文中，我们提出了一种联合 ESimCSE 无监督对比学会和 UDA 半监督学习大标签文本分

类新方法，并在该方法的有监督部分引入对抗训练技巧来提升模型的鲁棒性和泛化能力，我们使用一个业务数据集和一个公开数据集对方法有效性进行分析，我们分析了模型的准确率（Accuracy）、F1 值、人工验证准确率（MVA），并可视化了多个模型的输出向量化编码和两个数据集上不同模型的损失下降图，以及最终定义了一个对抗训练验证表，上述数据均表明了该方法的有效性。引入的 UDA 半监督训练通过数据增强技术来更好的对数据进行半监督学习；ESimCSE 模型通过学习输入的嵌入表示，将输入向量之间的特征变得更有却分性；对抗训练验证表的结果表明对抗训练可以让模型更能区分未见过（干扰过）的样本。实验表明三者结合后的最终模型整体的性能都有正向的提升，本研究对未来的相关领域的研究有着一定的启示和借鉴作用。

# 引用